# IMPRECISE MEANINGS AS A CAUSE OF UNCERTAINTY IN MEDICAL KNOWLEDGE-BASED SYSTEMS

*STEVEN J. HENKIND*

New York University,
Courant Institute of Mathematical Sciences, Department of Computer Science,
251 Mercer Street, New York, N.Y. 10012
and
Mount Sinai School of Medicine,
Department of Cardiology,
One Gustave L. Levy Place, New York, N.Y. 10029

## ABSTRACT

There has been a considerable amount of work on uncertainty in knowledge-based systems. This work has generally been concerned with uncertainty arising from the strength of inferences and the weighting of evidence. In this paper, we discuss another type of uncertainty: that which is due to imprecision in the underlying primitives used to represent the knowledge of the system. In particular, a given word may denote many similar but not identical entities. Such words are said to be lexically imprecise.

Lexical imprecision has caused widespread problems in many areas. Unless this phenomenon is recognized and appropriately handled, it can degrade the performance of knowledge-based systems. In particular, it can lead to difficulties with the user interface, and with the inferencing processes of these systems. Some techniques are suggested for coping with this phenomenon.

## INTRODUCTION: LEXICAL IMPRECISION

Specialized fields of knowledge can be viewed as having their own languages. These languages, which are subsets of natural language, are known as "sublanguages" [6]. Typically, sublanguages have a specialized vocabulary: for example, the vocabulary of medicine as found in a medical dictionary, or the vocabulary of law as found in a legal dictionary.

Since sublanguages are generally used in complex situations where there are difficult problems to be solved, and critical decisions to be made, it is important that the underlying vocabularies be well defined. In particular, the words need to have precise meanings. Clearly, if a given word does not have a precise meaning, then a sentence containing that word may be imprecise as well. If the sentence was intended to convey information relevant to the solution of a problem, then the imprecision might well lead to an incorrect solution.

Unfortunately, the vocabularies underlying most sublanguages are much less well defined than is commonly believed. For example, a word may be used to denote many similar, but not identical phenomena: In a recent review of the medical literature [4] we found more than a dozen definitions

129

for the important clinical phenomenon "pulsus paradoxus." All the definitions were intended to denote the same phenomenon. Yet they differed enough that two physicians, given the same patient, could reach opposite conclusions as to the presence or absence of a paradoxical pulse. We will say that words which have several similar, but not identical, meanings are "lexically imprecise."

The lexical imprecision to be found in medical terminology is not limited to certain exceptional words, but is, in fact, extremely common. For example, given that a patient has produced only 300 ml of rine during the past day, nearly every physician would state that the patient has oliguria (low urine output). But, if the the physicians were also told that the patient weighs 80 pounds, has received no fluids in the past 24 hours, and has been exercising heavily, then some would say that the patient is not oliguric, while others would continue to state that the patient has oliguria, albeit "to be expected under the circumstances." Oliguria is a term that is well-known to all physicians, but it is apparent that the use of the word is highly dependent upon who is using the word, under what circumstance, and so forth. A similar analysis can be performed for many other clinical entities, e.g., hypertension, etc.

It should be noted that we are not concerned here with words that are used to denote inherently imprecise entities. For example, the word "lethargic" must be somewhat imprecise because lethargy is very subjective in nature. However, we are extremely concerned with words that are used to denote ostensibly objective entities. The word "oliguria" can cause problems because although it is not well defined, nearly all physicians believe that it is. The word "lethargy," on the other hand, is much less likely to cause difficulty because physicians recognize that it is subjective.

Lexical imprecision is more than just a theoretical curiousity: it has, in fact, caused widespread difficulties in medicine and other fields. For example, "There are varying schools of thought among specialists in interpreting petit mal seizures. Some assign the designation *petit mal* to 3 percent of all forms of epilepsy; others classify 80 percent of seizures under this rubric .... In three recent papers, the results reported by the authors on a newly introduced anti-epileptic drug for the treatment of petit mal were respectively that it was highly effective, moderately effective, and ineffective. How much of this discrepancy is to be attributed to the drug or conditions of the trials, and how much to the different conditions regarded by the experiments as being petit mal?" [5].

## EFFECTS ON KNOWLEDGE-BASED SYSTEMS

One of the fundamental choices to be made in the construction of a knowledge-based system is the selection of an appropriate knowledge representation. Although rule-based representations are currently the most common choice, there are many other options, e.g., frames, semantic nets, scripts, etc.

At some point, however, all knowledge representations require a choice of semantic primitives. These primitives are the fundamental objects which a system will manipulate. In a knowledge-based system designed to perform medical diagnosis, for example, the semantic primitives would be various signs, symptoms, lab values, and diagnoses. Note that the semantic primitives are generally either lexical items (words), or numbers.

Consider a system which, when given some collection of signs and symptoms, deduces the patient's disease state. Call the set of all signs and symptoms $S$, the set of all subsets of $S$ (power set of $S$) $P(S)$, and the set of all diseases $D$. Then diagnosis can be viewed as a function from $P(S)$ to $D$; i.e., $F:P(S) \rightarrow D$. For the sake of argument, it is assumed here that each patient has one and only one disease. Furthermore, it is assumed that the diagnostic function is provided by a single domain expert.

Suppose, now, that a user makes a certain set of observations $O$ about a patient. In order to use the diagnostic system, the user will need to express these observations in terms of the semantic primitives of the system. In other words, $O$ must be mapped by the user into $P(S)$. Therefore, the computer-assisted diagnostic process is actually $O \rightarrow P(S) \rightarrow D$.

Unfortunately, due to lexical imprecision, two individuals may map the same observations differently. This can, in turn, lead to a different set of deductions by the system. For example, suppose that an expert creates the following system:

IF pulsus-paradoxus
> THEN tamponade-likely
> ELSE tamponade-not-likely

Also, suppose that user U1 defines pulsus paradoxus as an inspiratory decline in systolic arterial pressure of 10 mm Hg or more, and user U2 defines it as a decline of 13 or more. If a patient has an inspiratory decline of 12, then U1 would map his observation into the semantic primitive pulsus-paradoxus, but U2 would not. Therefore, the system would provide U1 with the conclusion that tamponade is likely, but would provide U2 with the opposite conclusion.

Certainly, both conclusions cannot be correct. The difference in conclusions is due to the fact that the map $O \rightarrow P(S)$ is not uniquely specified. In particular, this map varies from individual to individual, depending on the definitions that each person happens to use.

If a system is constructed by two or more domain experts, then lexical imprecision can lead to less than optimal performance of the diagnostic function $F:P(S) \rightarrow D$. Suppose, for example, that expert E1 provides the rule IF A THEN B, and expert E2 provides the rule IF B THEN C. Given A, the system will deduce C. But this may be incorrect if E1 and E2 have different definitions for the semantic primitive B. Of course, if a given expert is able to provide correct solutions to problems, then his knowledge must be, in some sense, internally consistent. The problem here, is that the components of his knowledge may be inconsistent with the components of another expert.

## LEXICAL IMPRECISION IS NOT LEXICAL AMBIGUITY

Many words have several distinct meanings. For example, the word "beat" can be used as a verb to denote the act of physically abusing someone, as a verb to denote the act of sailing a boat close to the wind, as a noun to denote a policeman's patrol area, and so forth. This phenomenon of multiple distinct meanings is referred to as "lexical ambiguity." Note that lexical ambiguity is not the same thing as lexical imprecision. A useful analogy is the following: If you open a dictionary and choose a word, it will have several distinct definitions. This is lexical ambiguity. If you open two dictionaries and choose the same word you will find sets of ver similar, but not identical definitions. This is lexical imprecision.

Lexical ambiguity can lead to difficulties in the processing of natural language by computer. From a syntactic standpoint, the possibility that a given word may have multiple meanings makes it necessar to select the correct part-of-speech for the word. For example, in parsing the sentence

131

"He is on his beat," it must be determined that "beat" is being used as a noun, and not as a verb. Since the meanings of a lexically imprecise word are all the same part-of-speech, lexical imprecision does not lead to problems in parsing.

Lexical ambiguity can also lead to difficulties in semantic analysis. Semantic analysis requires that the correct word-sense of an item be selected. For example, in analyzing the sentence fragment "Beat until you see the buoy," it must be determined that "beat" refers to the action of sailing close to the wind, and not to the infliction of physical violence. The dictionary of a typical natural language processing system may contain multiple definitions for a given word, but these definitions are invariably distinct. Hence, lexical imprecision has not posed many problems for semantic analysis.

It is at the level of pragmatics that lexical imprecision will cause the most difficulty for systems that use natural language. While subtle distinctions in meaning are of little consequence in the syntactic and semantic decomposition of sentences, these same distinctions can have profound consequences on a system that attempts to use those sentences.

There is much discussion in the Artificial Intelligence literature of techniques for handling lexical ambiguity. Birnbaum [2] provides a detailed review and analysis. On the other hand, there seems to be little, if any, discussion of methods for coping with lexical imprecision. The following section contains some preliminary suggestions.

## COPING WITH LEXICAL IMPRECISION

The ideal solution to the problem of lexical imprecision would be to eliminate it entirely. This would require that a precise set of definitions be established in each specialized field of knowledge -- presumably by a committee of experts. Furthermore, every individual in that field would need to agree upon and use those definitions. There has been, in fact, a great deal of effort in this direction, but with only mixed results. A notable success is the science of chemistry: once a molecule's structure has been determined, that substance has a name assigned to it that conveys the same meaning to every chemist. In the field of medicine, such efforts have been much less successful. For example, the American College of Cardiology, and the American Heart Association have published what they feel to be a standard definition for pulsus paradoxus [1], yet this definition was not to be found in any of the more than sixty papers in the literature that we surveyed [4].

Although these efforts to standardize terminology are extremely important, it will never be possible to completely eliminate lexical imprecision. As a field of knowledge expands, new discoveries are made, new measurement techniques devised, and entities are viewed at finer levels of granularity. This in turn renders some previously precise words less precise. For example, to characterize a patient as being "hypertensive" was actually quite precise a hundred years ago; but in modern medicine, such a characterization is far from adequate. Lexical imprecision is, and will remain, a ubiquitous phenomenon, and high performance knowledge-based systems will need to handle it in a reasonable fashion.

As was discussed previously, lexical imprecision can lead to difficulties both at the interface to a knowledge-based system, i.e., $O \rightarrow P(S)$, and within the system, i.e., $P(S) \rightarrow D$. There are several ways to lessen the impact at the interface level. One way is to eliminate that level as much as possible. In particular, the more observations that a system can make directly (rather than through a human intermediary), the less the damage that will be done due to individual differences in the mapping $O \rightarrow P(S)$. In some situations a system should be able to gather most of its input directly: for instance, monitoring a chemical plant. In many other situations, however, a human intermediary is essential. For example, computer vision and robotics notwithstanding, no machine is yet capable of performing a complete physical examination of a patient.

Problems at the interface level can also be decreased by insisting on quantification. For example, the question "Does patient have oliguria?" is highly susceptible to individual differences in the mapping $O \rightarrow P(S)$, but a request for the volume of urine output is much less problematical [3]. Unfortunately, not all phenomena can be quantified -- e.g., petit mal. Even for those phenomena which can be quantified, there are still potential difficulties to be aware of. For example, the normal ranges of various biomedical tests, e.g., enzyme assays, are not standardized, but actually vary from laboratory to laboratory.

Another consideration is that it must be certain that all observers are in fact measuring the same phenomenon (e.g., is pulsus paradoxus the inspiratory decrease in pulse pressure, or the inspiratory decrease in systolic pressure?). This could be encouraged as follows: when the system requests the value of an entity, it also provides a definition. For example, "Please input the measured value of pulsus paradoxus (inspiratory decline in systolic arterial pressure)." Of course, these definitions would need to be built into the system.

The incorporation of definitions into a system is also a potential solution to problems in the construction of the diagnostic function $P(S) \rightarrow D$. In particular, each expert could be encouraged to record his definitions for the semantic primitives with which he is working. Thus if expert E1 produces the rule A -> B, he would be expected to provide definitions for the primitives A and B. Before expert E2 entered the rule B -> C, he would be expected to check the definition of B, and so forth.

Unfortunately, it may be unreasonable to expect that definitions be provided for all the primitives of a large system. The demands in terms of increased development time and overhead could be enormous; nor is it clear how such definitions could be incorporated into existing systems. Furthermore, since the definitions of primitives are themselves composed from other primitives, it is impossible to enforce the complete consistency of the definitions within a system.

Fuzzy set theory [7] has some applicability as a tool to handle lexical imprecision. Many words are lexically imprecise because they are based upon cutoffs. For example, the lexical imprecision of "pulsus paradoxus" stems largely from the fact that different experts use different cutoffs, e.g., 10, 13, etc. These cutoffs in turn lead to discontinuous behavior, e.g., pulsus paradoxus is considered to be present given a cutoff of 10 but absent given a cutoff of 13. By modeling pulsus paradoxus as a fuzzy concept, the damaging effects of lexical imprecision could be greatly reduced because the presence or absence of pulsus would no longer be treated in a discontinuous fashion.

Note, however, that fuzzy techniques are not a solution for all cases of lexical imprecision. For example, fuzzy techniques could not reconcile the imprecision due to one observer measuring the decrease in pulse pressure, and another another observer measuring the decrease in systolic pressure.

## SUMMARY

Uncertainty is a major source of difficulty in the construction and use of knowledge-based systems. One type of uncertainty arises from the strength of the implication operator in inferences such as A -> B: e.g., if A then there is a 40% chance of B. Another type of uncertainty arises from the weighting of evidence: e.g., there is a 70% chance that the patient has A. In this paper, we have discussed another type of uncertainty -- that which is due to imprecision in the underlying primitives: e.g., two experts have a different conception of A.

Lexical imprecision can degrade the performance of knowledge-based systems. Effects can surface at both the user interface and inferencing levels.

Among the techniques to handle lexical imprecision are the direct acquisition of input data, quantification, the inclusion of definitions, and fuzzy set methods. Currently we are examining ways of incorporating these techniques into medical knowledge-based systems.

## ACKNOWLEDGMENTS

I would like to thank Ronald Yager, Malcolm Harrison, Ernie Davis, and Naomi Sager for their many helpful comments and suggestions.

## REFERENCES

[1] American College of Cardiology and American Heart Association. 1971. Glossary of cardiologic terms related to physical diagnosis: part IV. Arterial Pulses. *American Journal of Cardiology*, vol. 27, pp. 708-709.

[2] Birnbaum, L. 1985. Lexical ambiguity as a touchstone for theories of language analysis. *Proceedings of the Ninth IJCAI*, Los Angeles, California, pp. 815-820.

[3] Henkind, S., Benis, A., and Teichholz, L. 1986. Quantification as a means to increase the utility of nomenclature-classification systems. *Proceedings of the Fifth World Congress on Medical Informatics*, Washington D.C.

[4] Henkind, S., Benis, A., and Teichholz, L. 1985. The paradox of pulsus paradoxus. *Submitted for publication*.

[5] Kennedy, J., and Kossman, C. 1973. Nomenclatures in medicine. *Bulletin of the Medical Library Association*, vol. 61, pp. 238-252.

[6] Kittredege, R., and Lehrberger, J., eds. 1982. *Sublanguage: Studies of Language in Restricted Semantic Domains*, Walter de Gruyter, New York.

[7] Zadeh, L. 1965. Fuzzy Sets. *Information and Control* vol. 8, pp. 338-353.